\documentclass[letterpaper]{article} 
\usepackage{aaai23}  
\usepackage{times}  
\usepackage{helvet}  
\usepackage{courier}  
\usepackage[hyphens]{url}  
\usepackage{graphicx} 
\usepackage{natbib}  
\usepackage{caption} 
\usepackage{algorithm}
\usepackage{algorithmic}

\usepackage{newfloat}
\usepackage{listings}
\DeclareCaptionStyle{ruled}{labelfont=normalfont,labelsep=colon,strut=off} 
\floatstyle{ruled}
\newfloat{listing}{tb}{lst}{}
\floatname{listing}{Listing}
\usepackage[utf8]{inputenc} 
\usepackage{url}            
\usepackage{booktabs}       
\usepackage{amsfonts}       
\usepackage{nicefrac}       
\usepackage{microtype}      
\usepackage{xcolor}         
\usepackage{amsmath}        
\usepackage{algorithm}      
\usepackage{algorithmic}

\usepackage{graphicx}
\usepackage{multirow}
\usepackage{longtable}
\usepackage{enumitem, multirow, xspace}
\usepackage{caption}
\usepackage{subcaption}

\newcommand{\name}[0]{Pointerformer\xspace}

\newcommand{\etal}{\textit{et al}., }
\nocopyright

\title{Pointerformer: Deep Reinforced Multi-Pointer Transformer for the Traveling Salesman Problem}
\author{
    Yan Jin\textsuperscript{\rm 1},
    Yuandong Ding\textsuperscript{\rm 1},
    Xuanhao Pan\textsuperscript{\rm 1},
    Kun He\textsuperscript{\rm 1}\thanks{Corresponding author.},
    Li Zhao\textsuperscript{\rm 2},
    Tao Qin\textsuperscript{\rm 2},
    Lei Song\textsuperscript{\rm 2},
    Jiang Bian\textsuperscript{\rm 2}
}
\affiliations{
    \textsuperscript{\rm 1}School of Computer Science, Huazhong University of Science and Technology, China
    \textsuperscript{\rm 2}Microsoft Research Asia\\
     \{jinyan, yuandong, xhpan, brooklet60 \}@hust.edu.cn, \{lizo, taoqin, lei.song, jiang.bian\}@microsoft.com
}

\usepackage{bibentry}

\begin{document}

\maketitle

\begin{abstract}

Traveling Salesman Problem (TSP), as a classic routing optimization problem originally arising in the domain of transportation and logistics, has become a critical task in broader domains, such as manufacturing and biology. Recently, Deep Reinforcement Learning (DRL) has been increasingly employed to solve TSP due to its high inference efficiency. 
Nevertheless, most of existing end-to-end DRL algorithms only perform well on small TSP instances and can hardly generalize to large scale because of the drastically soaring memory consumption and computation time along with the enlarging problem scale. In this paper, we propose a novel end-to-end DRL approach, referred to as Pointerformer, based on multi-pointer Transformer. Particularly, Pointerformer adopts both reversible residual network in the encoder and multi-pointer network in the decoder to effectively contain memory consumption of the encoder-decoder architecture. To further improve the performance of TSP solutions, Pointerformer employs both a feature augmentation method to explore the symmetries of TSP at both training and inference stages as well as an enhanced context embedding approach to include more comprehensive context information in the query. Extensive experiments on a randomly generated benchmark and a public benchmark have shown that, while achieving comparative results on most small-scale TSP instances as SOTA DRL approaches do, Pointerformer can also well generalize to large-scale TSPs. 
\end{abstract}

\section{Introduction}

The Traveling Salesman Problem (TSP) is a well-known combinatorial optimization problem. It can be stated as follows: given a set of cities/nodes, a salesman departing from one city needs to traverse all other cities exactly once and finally returns to the start city. The objective of TSP is to find the shortest route for the salesman. In addition to its well-recognized theoretical importance as a classic combinatorial optimization problem, TSP also has a wide range of real-world applications, such as drilling of printed circuit boards~\cite{Alkaya2013}, X-Ray crystallography~\cite{Bland1989}, warehouse order picking~\cite{Madani2020}, transport routes optimization~\cite{Hacizade2018}, and many others~\cite{Matai2010}.

Due to both of its theoretical and practical importance, TSP has attracted a great number of research efforts in the past decades that attempted to address it using either exact or heuristic algorithms. In fact, the NP-hardness nature of TSP makes it computationally intractable to leverage exact algorithms to find the optimal solutions over a large-scale TSP, since the corresponding computation complexity increases exponentially with respect to the number of nodes. Hence, facing most real-world TSP applications, heuristic algorithms are usually adopted to obtain near-optimal solutions. However, to ensure achieving high-quality solutions, a few heuristic algorithms are designed to further rely on fine-tuned search strategies, which may significantly increase the time complexity for solving large-scale TSP.

Recently, there have been a soaring number of studies trying to solve TSP using either Supervised Learning (SL) or Reinforcement Learning (RL) approaches
~\cite{NIPS2015_29921001,Nowak2017,kool2018attention,Kwon2020,zheng2021combining,FuQZ21}. Compared with SL relying on the optimal solution as the learning labels, which is usually unknown when facing the large-scale TSP, RL yields the advantage since it can be applied to attain near-optimal solutions without requiring the existence of ground truth. Therefore, most recent studies tend to apply the Deep Reinforcement Learning (DRL) approach to solve the large-scale TSP.

Depending on the specific ways to construct solutions, DRL algorithms can be roughly divided into two main categories: search-based DRL~\cite{d2020learning,FuQZ21} and end-to-end DRL~\cite{kool2018attention,Kwon2020,kim2021learning}. By incorporating heuristic search operators with learning-based policy, search-based DRL can solve larger-scale TSP instances. However, they usually suffer from two major limitations. The first one lies in the inference efficiency issue, meaning that the search component usually takes a long time to terminate to obtain high quality solutions. Moreover, the obtained solution performance is very sensitive to the selection of search operators which is highly dependent on sophisticated domain knowledge. In contrast, end-to-end DRL algorithms are very efficient in generating solutions and bear much lower dependencies on domain knowledge. Therefore, end-to-end DRL algorithms are more suitable for many emerging TSP application scenarios, such as on-call routing~\cite{GHIANI20031} and ride hailing service~\cite{Xu2018}, that require to generate solutions in almost real-time. Nevertheless, most of existing end-to-end DRL algorithms only perform well on small TSP instances (no more than 100 nodes) and cannot generalize to larger instances easily. This is mainly due to the drastically soaring memory consumption and computation time along with the increasing nodes.

In this paper, we propose a novel scalable DRL method based on multi-pointer Transformer, denoted as \name, aiming to solve TSP in an end-to-end manner. While following the classical encoder-decoder architecture~\cite{vaswani2017attention}, this new approach adopts reversible residual network~\cite{gomez2017reversible, kitaev2019reformer} instead of the standard residual network in the encoder to significantly reduce memory consumption. Furthermore, instead of employing the memory-consuming self-attention module as in~\cite{kool2018attention,Kwon2020}, we propose a multi-pointer network in the decoder to sequentially generate the next node according to a given query. 
Besides addressing the issues of memory consumption, \name contains delicate design to further improve the model effectiveness. Particularly, to improve the effectiveness of obtained solutions, \name employs both a feature augmentation method to explore the symmetries of TSP at both training and inference stages as well as an enhanced context embedding approach to include more comprehensive context information in the query.

To demonstrate the effectiveness of \name, we conducted extensive experiments on two datasets, including randomly generated instances and widely used public benchmarks. Experimental results have shown that \name not only achieves comparative results on small-scale TSP instances as SOTA DRL approaches do, but also can generalize to large-scale TSPs. More importantly, while being trained on randomly generated instances, our approach can achieve much better performance on instances with different distributions, indicating a better generalization.


Our main contributions can be summarized as follows.
\begin{itemize}[topsep=.0in,leftmargin=0em,wide=0em,parsep=0em]
    \item We propose an effective end-to-end DRL algorithm without relying on any hand-crafted heuristic operators, which is the first end-to-end DRL approach that can scale to TSP instances with up to 500 nodes to the best of our knowledge. 
    \item Our algorithm applies an auto-regressive decoder with a proposed multi-pointer network to generate solutions sequentially without relying on any search components. Compared with existing search-based DRL algorithms, we can achieve comparable solutions while the inference time is reduced by almost an order of magnitude. 
    \item Besides scalability, we also show via extensive experiments that our approach can generalize well to instances that have varied distributions without re-training.
\end{itemize}

\section{Related work}

Here we highlight a few of the best traditional algorithms for solving TSP, and then focus on presenting the DRL algorithms that are more related to our work.

\textbf{Traditional TSP algorithms}. TSP is one of the most popular combinatorial optimization problems, and numerous algorithms have been proposed for TSP over the past decades. Traditional TSP algorithms can be classified into three categories, i.e., exact algorithms, approximate algorithms and heuristic algorithms. 
Concorde~\cite{applegate2007traveling} is one of the fastest exact solvers. It models TSP as a mixed-integer programming problem, and then adopts a branch and cut algorithm~\cite{padberg1991branch} to search the solution. Christofides \etal~\cite{christofides1976worst} proposed an approximation algorithm, and the approximation ratio of 1.5 is achieved by constructing the minimum spanning tree and the minimum perfect matching of the graph. LKH-3~\cite{helsgaun2017extension} is one of the SOTA heuristics, which uses the $k$-opt operator to search in the solution space, with the guidance of an $\alpha-$measure based on the minimum spanning tree. Among these traditional algorithms, the heuristics are the most widely used algorithms in practice, yet they are still time-consuming and difficult to be extended to other problems. 

Besides of these traditional algorithms, there are also works that attempted to utilize the power of machine learning and reinforcement learning techniques. Earlier machine learning approaches include the Hopfield neural network~\cite{hopfield_neural_1985} and self-organising feature maps~\cite{angeniol1988self}. There are several works like Ant-Q~\cite{gambardella1995ant} and Q-ACS~\cite{sun2001multiagent} that combined reinforcement learning with ant colony algorithm, and Liu and Zeng~\cite{liu2009study} used reinforcement learning to improve the mutation of a successful genetic algorithm called EAX-GA~\cite{nagata2006fast}. It is worth mentioning that a recent work, called VSR-LKH~\cite{zheng2021combining}, defined a novel Q-value based on reinforcement learning to replace the $\alpha-$value used by the LKH algorithm, and achieved a better performance on TSP.

\textbf{DL-based TSP algorithms}. DL-based TSP algorithms are mainly proposed in recent years, according to the way the solution is generated, they can be classified into two categories: end-to-end methods and search-based methods.

End-to-end methods create a solution from the scratch \cite{bello2016neural,dai_learning_2018,kim2021learning,kool2018attention,Kwon2020,nazari2018reinforcement,NIPS2015_29921001}. Vinyals \etal~\cite{NIPS2015_29921001} proposed a Pointer NetWork to solve TSP with SL. Bello \etal~\cite{bello2016neural} then used RL to train a PtrNet model to minimize the length of solutions. This method achieves better performance and has stronger generalization and scalability. To deal with both static and dynamic information, Nazari~\cite{nazari2018reinforcement} improved PtrNet, which is more effective than many traditional methods. Dai \etal~\cite{dai_learning_2018} proposed Structure2Vec which encodes partial solutions and predicts the next node. The Q-learning method is used to train the whole policy model. Attention Model in~\cite{kool2018attention} adopts the Transformer~\cite{vaswani2017attention} architecture and the model is trained through the REINFORCE algorithm with a greedy roll-out baseline. It shows the efficiency of Transformer in solving TSP. Then Kwon \etal proposed POMO~\cite{Kwon2020} using REINFORCE algorithm with a shared baseline. It leverages the existence of multiple optimal solutions of a combinatorial optimization problem. Currently, end-to-end methods perform well on TSP instances with nodes less than 100, but due to the complexity of the model and the low sampling efficiency of reinforcement learning, it is hard to extend them to a larger scale. 

Search-based methods start from a feasible solution and learn how to constantly improve it \cite{chen2019learning,d2020learning,fu2021generalize,joshi2019efficient,kool2022deep}. The improvement is often achieved by integrating with heuristic operators. For instance, Chen \etal proposed NeuRewriter~\cite{chen2019learning}, which rewrites local components through region-pick and rule-pick. They trained the model with Advantage Actor-Critic, and the reduced cost per iteration is used as its reward. Two approaches \cite{joshi2019efficient,kool2022deep} use SL to generate the heat maps of the given graphs, and then employ dynamic programming and beam search to find near-optimal solutions respectively. There is another method using Monte Carlo tree search (MCTS) to improve the solution such as Att-GCRN+MCTS~\cite{fu2021generalize}. They first trained a model to generate heat maps for guiding MCTS on small-scale instances by SL, based on which heat maps of larger TSP instances were then constructed by graph sampling, graph converting and heat maps merging. Finally, MCTS is used to search for solutions based on these heat maps. However, performance of such approaches highly depends on the number of iterations or search, which is usually time-consuming hindering their applications in time sensitive tasks.


\section{Problem formulation}

While there are many varieties of TSP problems, we focus on the classic two-dimensional Euclidean TSP in this paper. Let $G(V,E)$ denote an undirected full connection graph, where $V=\{v_i\mid 1\le i\le N\}$ represents all $N$ cities/nodes and $E=\{e_{ij}\mid 1\le i,j\le N\}$ is the set of all edges. Let $\mathit{cost}(i,j)$ be the cost of moving from $v_i$ to $v_j$, which equates the Euclidean distance between $v_i$ and $v_j$. We further assume $\mathit{depot}\in V$ denoting the depot city, from which the salesman starts the trip and will go back to it in the end. A route is defined as a sequence of cities. A route is feasible iff it starts from and ends at $\mathit{depot}$ while traverses all other cities exactly once.
Given a route $\tau$, its total cost, denoted by $L(\tau)$, can be calculated by Eq.~(\ref{eq:total_cost}), where $\tau_{[i]}$ denotes the $i$-th node on $\tau$ and $N=|\tau|$ is the length of $\tau$.

\vspace{-14pt}
\begin{equation}
\label{eq:total_cost}
L(\tau)=cost(\tau_{[N]},\tau_{[1]})+\sum_{i=1}^{N-1} cost(\tau_{[i]},\tau_{[i+1]}).
\end{equation}

A solution $\tau$ of TSP can be generated sequentially by selecting the next node from all nodes that are to be visited until returning to the $\mathit{depot}$. This can be seen as a Markov decision process. The decision of each step can be modeled by a deep neural network parameterized by $\theta$: $\pi_\theta(\tau_{[i]}\mid s,\tau[:i))$, where $s$ denotes a TSP instance and $\tau[:i)$ is the partial route on $\tau$ before the $i$-th step. The reward of each step is defined as the negative cost of the newly added edge. For each problem instance $s$, our goal is to maximize the expected cumulative reward defined as follows:
 \begin{equation}
    \label{objective_function}
    J(\theta\mid s)=\mathbb{E}_{\tau \sim p_{\theta}(\tau\mid s)} R(\tau),
 \end{equation}
 where $R(\tau)= -L(\tau)$ and $p_{\theta}(\tau\mid s)=\Pi_{i=1}^N\pi_\theta(\tau_{[i]}\mid s,\tau[:i))$.

According to the policy gradient theorem~\cite{sutton2000policy}, we can calculate the derivative of the objective function to update the model using many existing policy gradient algorithms.
\vspace{-3pt}
\begin{equation}
    \label{policy_gradient_theorem}
    \nabla_{\theta} J(\theta\mid s)=\mathbb{E}_{p_\theta(\tau\mid s)}\left[\nabla_\theta \log p_\theta(\tau\mid s) R(\tau)\right]
\end{equation}



\section{The \name approach}

\begin{figure*}[!htp] 
    \centering 
    \includegraphics[width=0.6\textwidth]{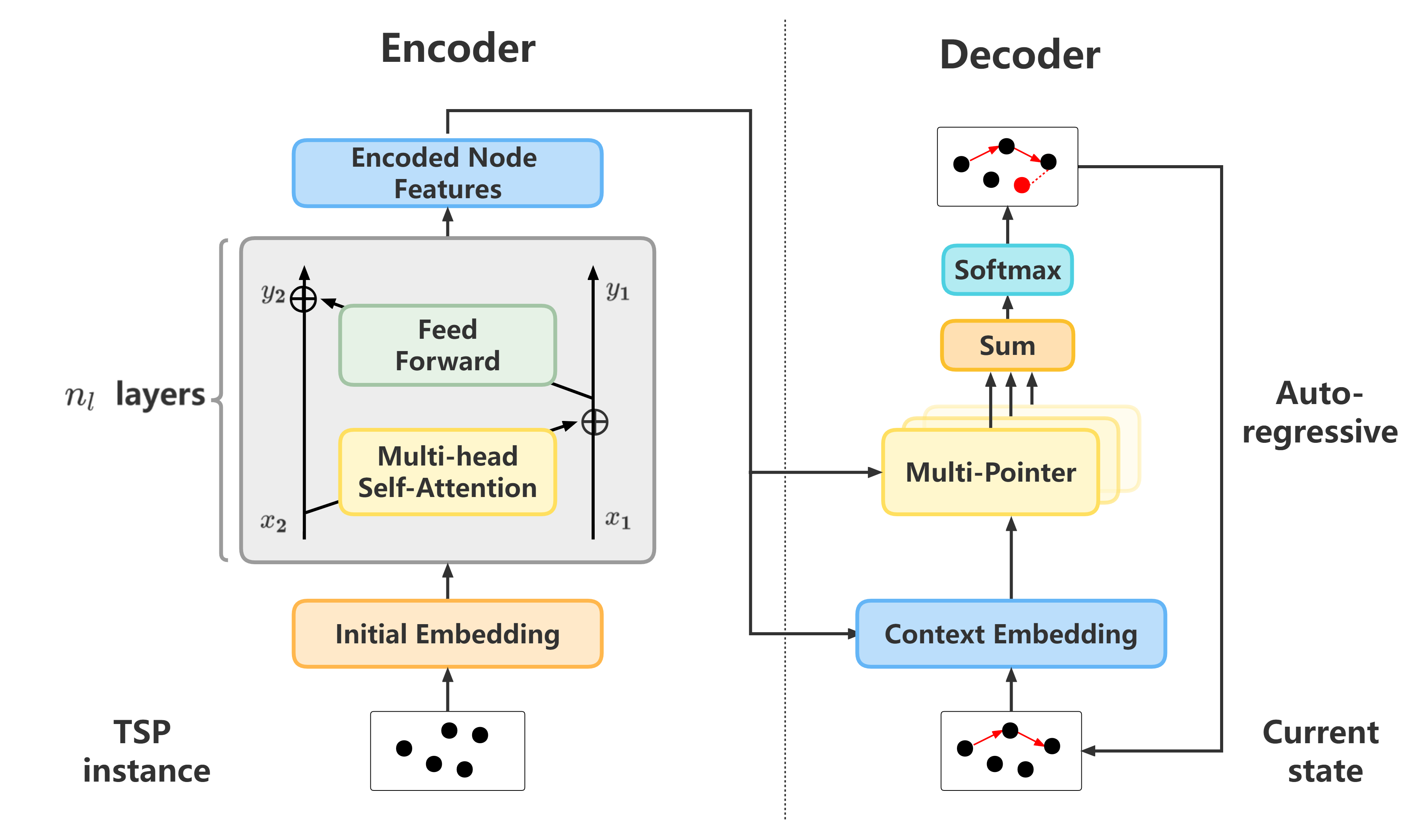} 
    \caption{The overall architecture of \name. First, multiple attention layers are applied to encode the nodes of the input TSP instance. Next, a multi-pointer network is used to sequentially decode the solution by a query composed of an enhanced embedding.} 
    \vspace{-17pt}
    \label{Fig.architecture} 
\end{figure*}

The proposed \name is an end-to-end DRL algorithm based on multi-pointer transformer which combines a transformer encoder and an auto-regressive decoder. The general framework of \name is illustrated in Figure~\ref{Fig.architecture}. 


In principle, \name applies multiple attention layers that consist of multi-head self-attention and feed-forward layers to encode the input nodes for obtaining an embedding of each node. Then, a multi-pointer network with a single head attention is employed to decode sequentially according to a query composed of an enhanced context embedding. Here, the enhanced context embedding contains not only information about the instance itself and nodes that are to be visited, but also information about nodes that have been visited. The solution is generated by choosing a node at each step according to the probability distribution given by the decoder, where all visited nodes are masked so that their probability is 0. Finally, the proposed \name is trained with a modified REINFORCE algorithm, which is based on a shared baseline for policy gradients while unifying the mean and variance of a batch of instances. In the following subsections, we describe the key components of \name. 

\subsection{Reversible residual network based encoder}
The encoder is an important ingredient for the \name architecture. As we mentioned before, the resource consumed by the origin Transformer~\cite{vaswani2017attention} increases dramatically as the length of the input sequence increases, which equates the number of nodes in TSP. Therefore, we adopt a Transformer without positional encoding but including a reversible residual network, in order to scale to large TSP instances. To our knowledge, the reversible residual network has not been introduced into the DRL approaches of combinatorial optimization problems before.

In the classic two-dimensional Euclidean TSP setting, each node is denoted by its coordinates $(x, y)$ and nothing else. To obtain a robust embedding for each node, we propose a feature augmentation mechanism such that each node is denoted by $(x, y, \eta)$, where $\eta=atanh\frac{y}{x}$. Furthermore, inspired by the data augmentation in POMO~\cite{Kwon2020} that generates 8 equivalent instances of each instance by flipping and rotating its underlying graph, we finally use them on the defined feature to obtain 24 features for each node. These features will be input of the initial embedding layer.

After the initial embedding layer, nodes will go through the encoder with multiple residual layers, each of which is constituted by a multi-head self-attention ($\mathbf{MHA}$) sub-layer and a feed-forward ($\mathbf{FF}$) sub-layer. Here, we employ the reversible residual network~\cite{gomez2017reversible, kitaev2019reformer} to save memory consumption. Different from residual networks where activation values of all residual layers need to be stored in order to calculate the derivations during back-propagation, in reversible residual networks, $\mathbf{MHA}$ and $\mathbf{FF}$ maintain a pair of input and output embedding features $(X_1, X_2)$ and $(Y_1, Y_2)$ so that derivations can be calculated directly. Below we illustrate the details in Eq.~(\ref{rev_forward}) and (\ref{rev_backward}):
    \begin{equation}
    \label{rev_forward}
    \begin{aligned}
        Y_1&=X_1+\mathbf{MHA}(X_2), \\
        Y_2&=X_2+\mathbf{FF}(Y_1). \\
    \end{aligned}
    \end{equation}
Obviously, the input embedding features $(X_1, X_2)$ can be calculated from the output embeddings $(Y_1, Y_2)$ easily during back-propagation: 
    \begin{equation}
        \label{rev_backward}
    \begin{aligned}
        X_2&=Y_2-\mathbf{FF}(Y_1), \\
        X_1&=Y_1-\mathbf{MHA}(X_2). \\
    \end{aligned}
    \end{equation}
    
Note that the deeper of the residual network, the more memory the reversible residual network can save. In our work, we apply $\mathbf{MHA}$ and $\mathbf{FF}$ of six layers, we can observe dramatic reduction of memory consumption without affecting performance.

\subsection{Multi-pointer network based decoder}
The decoder is an auto-regressive process that is to sequentially generate a feasible route for each TSP instance. A context embedding is used to represent the current state, and is used as a query to interact with embeddings of nodes that are to be selected. The context embedding is updated constantly as more nodes are selected until a feasible route is obtained. The auto-regressive decoder is generally very fast but memory-consuming, mainly due to the attention module used in the query. To alleviate this, we improve our decoder by integrating the following distinguishing features.

\textbf{Enhanced Context Embedding}. Recall that a route $\tau$ of TSP is composed of a sequence of nodes on it. We propose an effective and enhanced context embedding that contains the following information $h_{\tau_{[1]}}, h_{\tau_{[t]}}, h_g$, and $h_\tau$, where $t=|\tau|$ is used to denote the length of $\tau$: 
\begin{itemize}[topsep=.0in,leftmargin=0em,wide=0em,parsep=0em]
    \item $h_{\tau_{[1]}}$ embedding of the first node on $\tau$: A static information that is the embedding of $\mathit{depot}$;
    \item $h_{\tau_{[t]}}$ embedding of the last node on $\tau$: A dynamic information that is updated according to the current route;
    \item $h_g$ graph embedding: To encode the whole TSP instance, which is the summation of embeddings of all nodes in the instance: $h_g=\sum_{i=1}^N h^{enc}_i$, where $h^{enc}_i$ is the embedding of the $i-$th node obtained by the encoder;
    \item $h_\tau$ embedding of $\tau$: To encode the current partial route, which is the summation of embeddings of all nodes on $\tau$ $h_\tau=\sum_{i=1}^{t}h_{\tau_{[i]}}^{enc}$.
\end{itemize}


The enhanced context embedding is used as a query $q_t$, which is computed by $q_t=\frac{1}{N}(h_{g}+h_{\tau})+h_{\tau_{[t-1]}}+h_{\tau_{[1]}}$. 
Since the graph embedding is able to reflect different graph structures while information about $\mathit{depot}$ and the last visited node is crucial for selecting future nodes, we include such information to guide the decoder similar as in the previous DRL algorithms~\cite{kool2018attention,Kwon2020}. Additionally, we also utilize $h_\tau$ in our decoder which is ignored in previous solutions. The motivation is that even with the same first and last nodes, two routes may cause different distributions over nodes that are to be visited. As shown in our experiments, such information is crucial, particularly for instances from practical applications. 
Notice that we normalize the graph embedding and the current partial route embedding by dividing the total number of nodes $N$.

    
    \textbf{A Multi-pointer Network}. At each step, the above enhanced context embedding is used to interact with all nodes that are to be visited to output a probability distribution over them. We devise a multi-pointer network to better utilize the context embedding. More specifically, we linearly project the queries $q_t$ and keys $k_j$ (embedding of the $j-$th node given by the encoder) to $d_k$ dimensions by using $H$ different linear projections for each of them. For each projection, we are able to obtain an interaction between the query and node $j$ via a dot operator and normalization by $\sqrt{d_k}$. The final interaction is simply evaluated by an average operator over all $H$ interactions, namely, $\mathit{PN} = \frac{1}{H}\sum_{h=0}^{H}{\frac{{(\mathbf{q}_{t}W_h^q})^{T} (\mathbf{k}_{j}W_h^k)}{\sqrt{d_{\mathrm{k}}}}}.$
    
    
    
    We further minus $\mathit{PN}$ by the cost between the last node $i$ of the partial route and node $j$ to obtain the interaction score between $i$ and $j$: $score_{ij} = \mathit{PN} - \mathit{cost}(i,j)$. By doing so, we encourage the approach to start from a good policy that is always selecting the nearest node as the next one to visit. Comparing to starting from a random policy, this will accelerate our training procedure considerably. 
    


    Similar to \cite{bello2016neural}, the probability is obtained by Eq.~(\ref{eq:clip}), where we clip the score with $\mathit{tanh}$ and mask all visited nodes. Here, $C$ is a coefficient that controls the range of values. The larger $C$ is, the smaller of the entropy, hence can be seen as a parameter to control the trade-off between exploitation and exploration during training. We will show via ablation experiments that the value of $C$ has a significant impact on performance.
    
    \begin{small}
    \begin{equation}
    \vspace{-5pt}
    u_{ij}  = \begin{cases}C \cdot \tanh \left(score_{ij}\right) & \text{node $j$ is to be visited}\\ -\infty & \text { otherwise }\end{cases}
    \label{eq:clip}
    \end{equation}
    \end{small}
    
    Finally, we are able to compute the output probability vector $p$ using a $\mathit{softmax}$ function. 
    

\subsection{A modified REINFORCE algorithm}
We train our \name model by using the REINFORCE algorithm~\cite{williams1992simple}, whose baseline applies diverse greedy roll-outs of all instances for policy gradient. Inspired by POMO~\cite{Kwon2020}, our decoder also starts from $N$ different nodes for each TSP instance with $N$ nodes. By taking each node as the depot, for each TSP instance $i$, 
we can sample $N$ feasible routes $\tau_i = \left\{\tau_i^1, \tau_i^2, \ldots, \tau_i^N\right\}$ by Monte Carlo sampling method. 
Therefore, given a batch containing $B$ TSP instances, we can obtain $B\times N$ routes, which can be used to train our policy according to Eq.~(\ref{policy_gradient_theorem}). However, directly applying REINFORCE will cause the algorithm hard to converge because of high variance of costs among different instances. In order to alleviate such a problem, we further use a variance-consistent normalization mechanism before training, which can increase the speed of convergence while also stabilizes the training. More details can be found in Eq.~(\ref{eq:return_gradient}), 
 where $\mu(\tau_i)$ and $\sigma(\tau_i)$ are the mean and variance of the $N$ trajectories of instance $i$, respectively. One can easily observe that $\frac{R\left(\tau^{j}_i\right)-\mu(\tau_i)}{\sigma(\tau_i)}$ is an unbiased estimation of the TSP objective function, which eliminates the effect of different rewards among different instances.




\begin{small}
\begin{equation}
\begin{aligned} 
\label{eq:return_gradient}
    \nabla_{\theta} J(\theta) &\approx \frac{1}{B\times N} \sum_{i=1}^{B}\sum_{j=1}^{N}\left(\frac{R\left(\tau^{j}_i\right)-\mu(\tau_i)}{\sigma(\tau_i)}\right) \nabla_{\theta} \log p_{\theta}\left(\tau^{j}_i \mid s\right), \\
   \mu(\tau_i)&=\frac{1}{N} \sum_{j=1}^{N} R\left(\tau_i^{j}\right), \\
   \sigma(\tau_i)&=\frac{1}{N} \sum_{j=1}^{N} \left( R\left(\tau_i^{j}\right)-\mu(\tau_i)\right)^2.
\end{aligned} 
\end{equation}
\end{small}

\section{Experiment}

To evaluate the efficiency of \name, we compare its performance with SOTA DRL approaches. We train and test \name on randomly generated instances, and verified its generalization on a public benchmark.

\subsection{Benchmark Instances}
\begin{itemize}[topsep=.0in,leftmargin=0em,wide=0em,parsep=0em]
    \item \textbf{TSP\_random}: Uniformly sample a certain number of nodes from the unit square of $[0,1]^2$. It includes five sets of TSP instances with $N$ = 20, 50, 100, 200, 500. Same as in Att-GCRN+MCTS~\cite{fu2021generalize}, for TSP instances with $N\le 100$, we sample 10,000 instances for each set, while for larger instance with $N \ge 200$, the set size is 128. The same benchmark is also widely adopted to testify existing DRL approaches except that they only consider instances with $N \le 100$; 
    \item \textbf{TSPLIB}: A well-known TSP library~\cite{reinelt1991tsplib} that contains 100 instances with various nodes distributions. These instances come from practical applications with size ranging from 14 to 85,900. In our experiment, we consider all instances with no more than 1002 nodes.

\end{itemize}

\subsection{Baselines}
The following SOTA DRL algorithms are considered as our baselines.

\textbf{End-to-end DRL algorithms}:
\begin{itemize}[topsep=.0in,leftmargin=0em,wide=0em,parsep=0em]
    \item AM \cite{kool2018attention}: A model based on attention layer is trained using the REINFORCE algorithm with a deterministic greedy roll-out baseline. AM can achieve good performance on small-scale TSP instances;
    \item POMO \cite{Kwon2020}: To reduce the variance of advantage estimation, POMO improves the algorithm in AM such that it generates $N$ trajectories for each instance with $N$ nodes and uses data augmentation to improve the quality of solutions during validation;
    \item AM+LCP \cite{kim2021learning}: It proposes a training paradigm for solving TSP called termed learning collaborative policy. It distinguishes policy seeder and policy reviser, which focus on exploration and exploitation, respectively. 
\end{itemize}

\textbf{Search-based DRL algorithms}:

\begin{itemize}[topsep=.0in,leftmargin=0em,wide=0em,parsep=0em] 
    \item DRL+2opt \cite{d2020learning}: DRL+2opt guides the search of 2-opt operator through DRL. The combination of reinforcement learning and heuristic search operator constantly improve solutions to achieve good results.
    \item Att-GCN+MCTS \cite{FuQZ21}: It trains a model to generate heat maps for guiding MCTS on small-scale instances by supervised learning, based on which heat maps of larger instances are then constructed by graph sampling, graph converting and heat maps merging. Finally, MCTS is used to search for solutions based on the heat maps.
\end{itemize}
 
\subsection{Hyper-parameters}
 In our experiments, we only use instances from \textbf{TSP\_random} to train various models corresponding to instances with different nodes. During each training epoch, 100,000 instances are randomly sampled. To train models for instances of size $N\le 200$, we use a single GPU V100 (16G) with batch size $B=64$, while for other cases the models are trained on four GPUs V100 (32G) with batch size $B=32$. Adam is used as the optimizer for all models with a learning rate $\eta=10^{-4}$ and a weight decay $\omega=10^{-6}$. We use 6 layers in the encoder ($n_t = 6$) and let $d_k=128$ and $H=8$ of multi-pointer in the decoder. The number of heads is 8 in the $\mathbf{MHA}$ layer. When evaluating on \textbf{TSP\_random}, the batch size $B$ is 128 for instances with $N\le 200$, while $B=64$ for other cases. Our algorithm is implemented based on PyTorch~\cite{NEURIPS2019_9015} and the trained models and the related data will be publicly available once the paper is accepted \footnote{https://github.com/Pointerformer/Pointerformer}.

\subsection{Experimental Results}

\begin{table*}[htb]
    \centering
    \caption{Comparison results on instances from \textbf{TSP\_random}.}
    \vspace{-10pt}
    \label{tab:TSP_random1}
    \resizebox{\textwidth}{!}{
    \begin{tabular}{c|ccc|ccc|ccc|ccc|ccc}
        \toprule[1pt]
        Method & \multicolumn{3}{c|}{TSP\_random20} & \multicolumn{3}{c|}{TSP\_random50} & \multicolumn{3}{c|}{TSP\_random100} & \multicolumn{3}{c|}{TSP\_random200} & \multicolumn{3}{c}{TSP\_random500} \\ 
        & Len & Gap & Time &Len & Gap & Time & Len & Gap & Time & Len & Gap & Time & Len & Gap & Time \\
        &  & (\%) &  & & (\%) &  &  & (\%) &  & & (\%) &  &  & (\%) &  \\
         \hline
         OPT &3.83 &  & & 5.69 & & & 7.76 &  &  & 10.72 &  &  & 16.55 &  & \\
         \hline
         AM & 3.83   & 0.06\  & 5.22s & 5.72  & 0.49\  & 12.76m & 7.94 & 23.20\  & 32.72m & - & - & - & - & - & - \\
         POMO & 3.83  & 0.00\  & 36.86s & 5.69 & 0.02\  & 1.15m & 7.77 & 0.16\  & 2.17m & - & - & - & - & - & - \\
         AM+LCP & 3.84  & 0.00\  & 30.00m & 5.70 & 0.02\  & 6.89h & 7.81 & 0.54\  & 11.94h & - & - & - & - & - & - \\
         DRL+2opt & 3.83 & 0.00\  & 3.33h & 5.70 & 0.12\  & 4.62m & 7.82 & 0.78\  & 6.57h & - & - & - & - & - & - \\
         Att-GCN+MCTS & 3.83 & 0.00\  & 1.6m & 5.69 & 0.01\  & 7.90m & 7.76 & 0.04\  & 15m & 10.81 & 0.88\  & 2.5m & 16.97 & 2.54\  & 5.9m \\
        \hline
         \textbf{\name} & \textbf{3.83} & \textbf{0.00} & 5.82s & \textbf{5.69} & 0.02\  & 11.63s & 7.77 & 0.16\  & 52.34s & \textbf{10.79} & \textbf{0.68\ } & \textbf{5.54s} & 17.14 & 3.56\  & 59.35s \\ 
        \bottomrule[1pt]
        \end{tabular}
    }
    \begin{minipage}{\textwidth}
    \small
    TSP20, TSP50 and TSP100: 10,000 instances; TSP200 and TSP500: 128 instances.
    \end{minipage}
\end{table*}

\begin{table*}[htb]
    \centering
    \caption{Comparison results on practical instances from \textbf{TSPLIB}.}
    \vspace{-10pt}
    \label{tab:tsplib}
    \resizebox{\textwidth}{!}{
    \begin{tabular}{c|ccc|ccc|ccc}
        \toprule[1pt]
        Method  & \multicolumn{3}{c|}{TSPLIB1$\sim$100} & \multicolumn{3}{c|}{TSPLIB101$\sim$500} & \multicolumn{3}{c}{TSP501$\sim$1002} \\ 
         & Len & Gap & Time & Len & Gap & Time & Len & Gap & Time \\
		 &  & (\%) &  &  & (\%) &  &  & (\%) &  \\
         \hline
         OPT & 19454.17 &    &   & 40842.43 &  &   & 62427.71 &  &   \\
         \hline
         AM & 22283.67 & 15.36\  & 0.23s & 72137.93 & 78.18\  & 0.86s & 140664.29 & 139.02\  & 5.79s \\
         POMO & \textbf{19628.67} & 1.20\  & 1.41s & \underline{43652.77} & 6.99\  & 1.55s & 82162.29 & 26.93\  & 3.49s \\
         DRL+2opt & 19916.50 & 2.43\  & 15.20m & 46651.40 & 13.85\  & 27.92m & 82797.71 & 42.57\  & 1.24h \\
         \hline
         \textbf{\name(Model100)}  & \underline{19728.50} & 1.33\  & 0.20s & \textbf{42963.20} & 5.43\  & 0.46s & \underline{75081.43} & 18.65\  & 5.14s\\
         
         \textbf{\name(Model200)}  & 20135.00 & 2.91\  & 0.20s & 43810.67 & 8.37\  & 0.46s & \textbf{73915.57} & 18.20\  & 5.14s\\
      \toprule[1pt]
    \end{tabular}
    }
    \vspace{-17pt}
\end{table*}

To show effectiveness of \name, we first train models with different number of nodes, denoted by \textbf{Model$N$} with $N=$ 20, 50, 100, 200, and 500, respectively. For training \textbf{Model$N$}, random instances of size $N$ are sampled from \textbf{TSP\_random} using parameters as stated in the above section.

\begin{figure}[!htp] 
    \centering
        \includegraphics[width=0.7\linewidth]{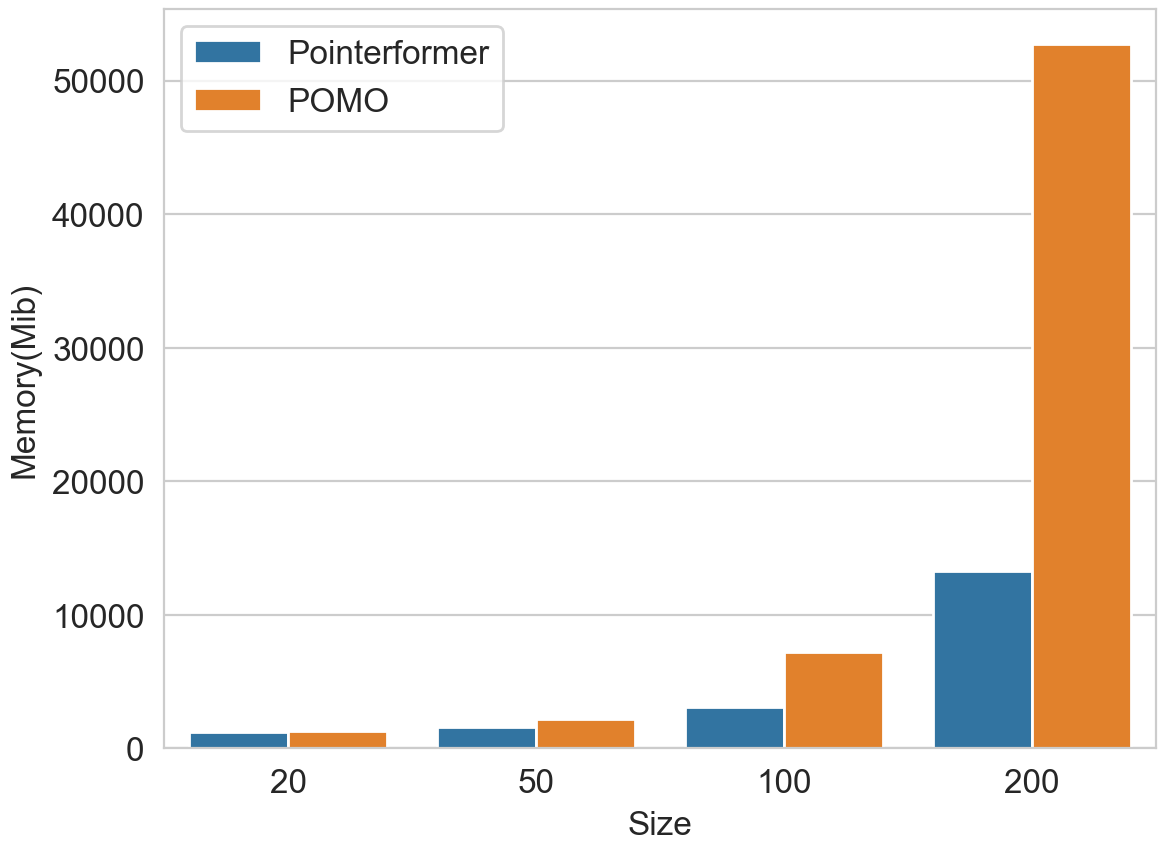} 
        \caption{Comparison of memory consumption between \name and POMO. Along with the enlarging problem size, the memory consumption of POMO increases sharply, while our model increases gradually.}
        \vspace{-17pt}
        \label{Fig:memory_time}
    \end{figure}

We have conducted the experiment on \textbf{TSP\_random} and a further study of generalization on \textbf{TSPLIB}, in all of which we observe advantages of \name over others. For the group of \textbf{TSP\_random} benchmark, the results are shown in Table~\ref{tab:TSP_random1}, from which we can see that \name has the best trade-off between efficiency and optimality compared to others. \name can achieve results of relatively small gaps to the optimal solutions, denoted by OPT and achieved by running the exact algorithm Concorde. More importantly, one easily observes that \name can scale to TSP instances with up to 500 nodes while other DRL algorithms except Att-GCN+MCTS quickly run out of memory for TSP instances with $N>100$ (indicated by - in Table \ref{tab:TSP_random1}). In Fig.~\ref{Fig:memory_time}, we also compare memory consumption of our model with the SOTA DRL approach POMO trained on instances of different size. One easily observes that along with the enlarging problem size, the memory consumption of POMO increases sharply, while our model increases gradually. Note that since the architecture of POMO is the most similar with ours, so it is more fair to use POMO for comparison of memory consumption when comparing to other DRL models. Comparing to search-based approach, the solutions obtained by \name may be slightly worse than Att-GCN+MCTS on TSP instances with 500 nodes. However, we can accelerate the computing time by up to 6 times (5.9m to 59.35s). In particular, we can attain a better result on TSP instances with 200 nodes in less time. We shall mention that results of Att-GCN+MCTS are taken directly from~\cite{fu2021generalize}, where the search component is implemented in C++ and runs in a CPU with 8 cores in parallel.  
 
    To the best of our knowledge, \name is the first end-to-end DRL algorithm that can scale to TSP instances with more than 100 nodes while still achieve comparable results as search-based DRL approaches, but in shorter time.

  In order to evaluate the generalization of the proposed \name, we apply \textbf{Model100} directly to the \textbf{TSPLIB} instances, similar for the baseline algorithms AM, POMO, and DRL+2opt. Note that we do not compare with Att-GCN+MCTS and AM+LCP here, since we have not figured out how to extend Att-GCN+MCTS to non-random setting, while the implementation of AM+LCP is not publicly available. To further verify the importance of scalability, we also apply \textbf{Model200} to these instances, which are unavailable for the baselines due to lack of scalability. We shall see that \textbf{Model200} has better generalization comparing to \textbf{Model100}, particularly for large-scale instances. Table~\ref{tab:tsplib} summarizes the results of \name in comparison with the three baselines on instances from \textbf{TSPLIB}, where we classify instances in \textbf{TSPLIB} into three groups according to their sizes, i.e., TSPLIB1$\sim$100, TSPLIB101$\sim$500, and TSPLIB501$\sim$1002. From the results, we can see that POMO performs the best on instances with no more than 100 nodes and the second best on instances between 101 to 500 nodes. While  \name (\textbf{Model100}) performs the best on instances between 101 to 500 nodes and the second best on the other two groups. One notices that most instances of second group are around 100 nodes, so \name (\textbf{Model100}) has the best performance and POMO has the second best performance for them. \name (\textbf{Model200}) and \name (\textbf{Model100}) perform the best and the second best on instances with more than 500 nodes, indicating that our model generalizes best to large-scale instances.

	


Due to space limit, we only show partial results in the paper. We shall refer interested readers to the appendix for a complete list of our experiment results. 

\subsection{Ablation experiments}
In this section, we present some ablation experiment results that explain some important choices of our approach.

\textbf{Importance of different components of \name}
To assess the influence of some key components to the performance of \name, we carry out an additional ablation study to compare \name and its four variants on instances from \textbf{TSP\_random} with 200 nodes (TSP\_random200). The results are summarized in Table~\ref{tab:ablations}. The first variant only uses the coordinates of each node as inputs without any feature augmentation (denoted by w.o. feature augmentation in the table). The second variant removes the embedding of the current partial route from the context embedding (denoted by w.o. enhanced context embedding). 

\begin{table}[!htb]
    \centering
    \caption{Ablations of three key elements of \name on TSP\_random200.}
    \vspace{-2pt}
    \begin{tabular}{c|c|c}
        \bottomrule
        Algorithm & Len & Gap\\
        \hline
        \name & 10.793 & 0.68\% \\
        w.o. feature augmentation & 10.813 & 0.87\% \\
        w.o. enhanced context embedding & 11.013 & 2.73\% \\
        w.o. multi-pointer network & 10.797 & 0.72\% \\
        \bottomrule
    \end{tabular}
    \vspace{-17pt}
    \label{tab:ablations}
\end{table}

\begin{figure}[!htb] 
    \centering 
    \vspace{-10pt}
    \includegraphics[width=0.4\textwidth]{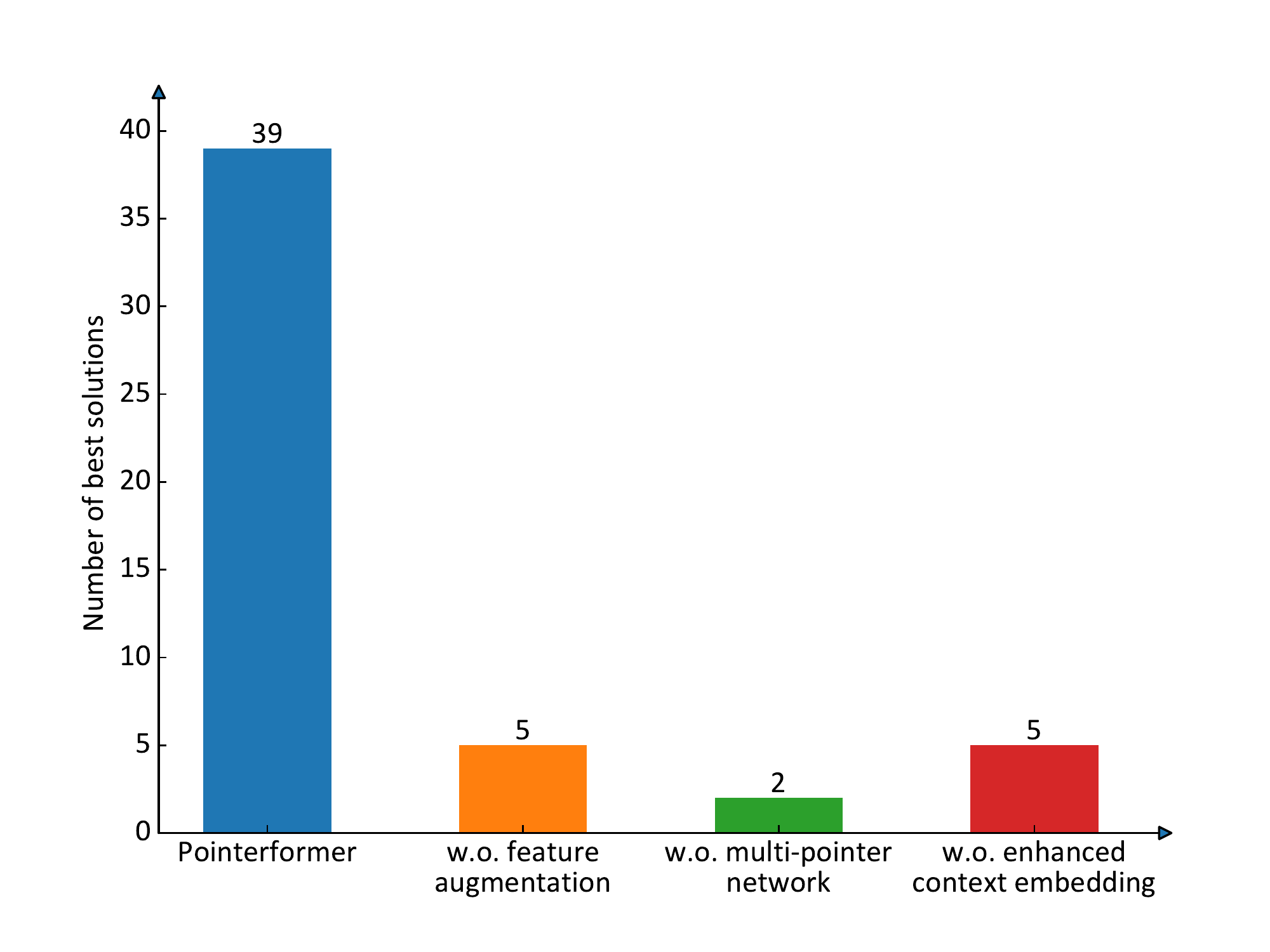}
    \caption{Ablations of three key elements of \name on TSP\_random200.}
    \vspace{-10pt}
    \label{Fig:ablation} 
\end{figure}

And the third variant does not use the multi-pointer networks, denoted by w.o. multi-pointer network. From Table \ref{tab:ablations}, it is clear that \name achieves the best performance comparing to all the variants, which indicates all components play positive roles to our algorithm. Furthermore, we apply these models directly test the instances from \textbf{TSPLIB} and provide their comparisons in Figure \ref{Fig:ablation}. \name with all components outperforms the three variants, indicating that these components are also important for the generalization of \name.

\begin{figure}[H] 
    \centering 
    \subfloat[TSP\_random100]{\includegraphics[width=0.25\textwidth]{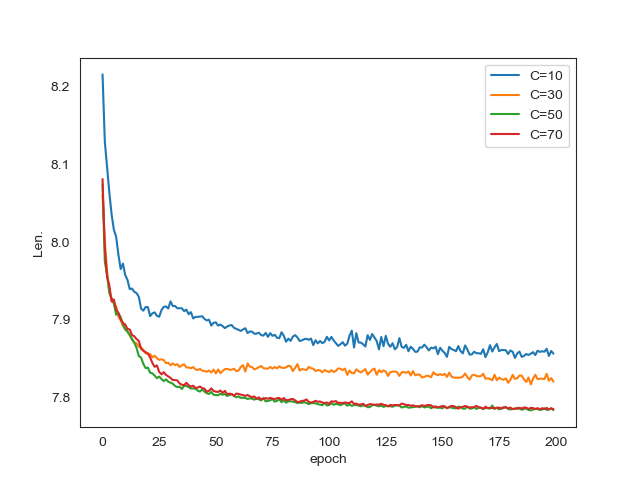}}
	\subfloat[TSP\_random200]{\includegraphics[width=0.25\textwidth]{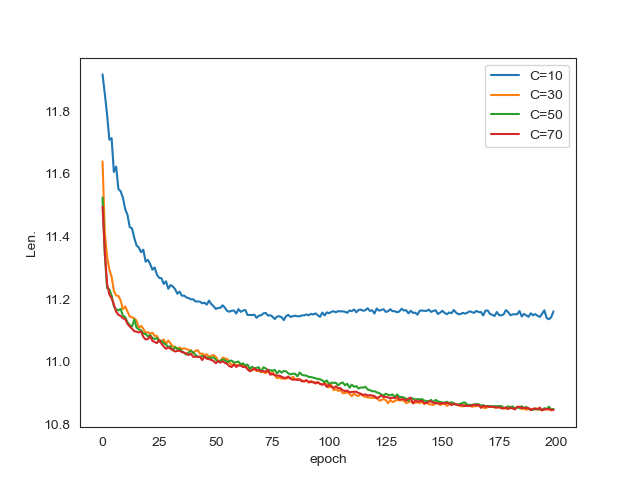}}\quad
    \caption{Comparison of different clipping values. } 
    \label{Fig.comparison} 
\end{figure}

\textbf{How $C$ affects the performance} As we mentioned before, the value of $C$ in Eq.~(\ref{eq:clip}) affects the entropy of distributions over nodes that are to be visited, which consequently has a big impact on the trade-off between exploitation and exploration. As shown in Figure~\ref{Fig.comparison}, once $C$ is too small ($\le 10$ in our experiment on instances from \textbf{TSP\_random} with 100 and 200 nodes), namely, it encourages exploitation too much, then the algorithm will quickly fall into a local optima. However, once $C$ is big enough ($\ge 50$), the difference will be negligible and the algorithm will constantly converge to a better policy than the one for $C=10$.

\section{Conclusion}

In this paper, we propose an end-to-end DRL approach called \name to solve large-scale TSPs. By integrating feature augmentation, reversible residual network, and enhanced context embedding with the well-known Transformer architecture, \name can achieve comparable results as SOTA algorithms do but using less resources (time or memory). While being memory-efficient, \name can be scaled to handle TSP instances with 500 nodes, 
that existing end-to-end DRL approaches could not solve. 
More importantly, we show via extensive experiments on well-known TSP instances with different distributions that our approach has better generalization.
For future work, we will explore how to extend our approach to address the more complicated problem of vehicle routing and other combinatorial optimization problems.

\bibliography{ref}

\newpage
\section{Appendix}

\subsection{More results}
\begin{figure}[!htb]
	\centering
	\subfloat[TSP\_random500]{\includegraphics[width=0.225\textwidth]{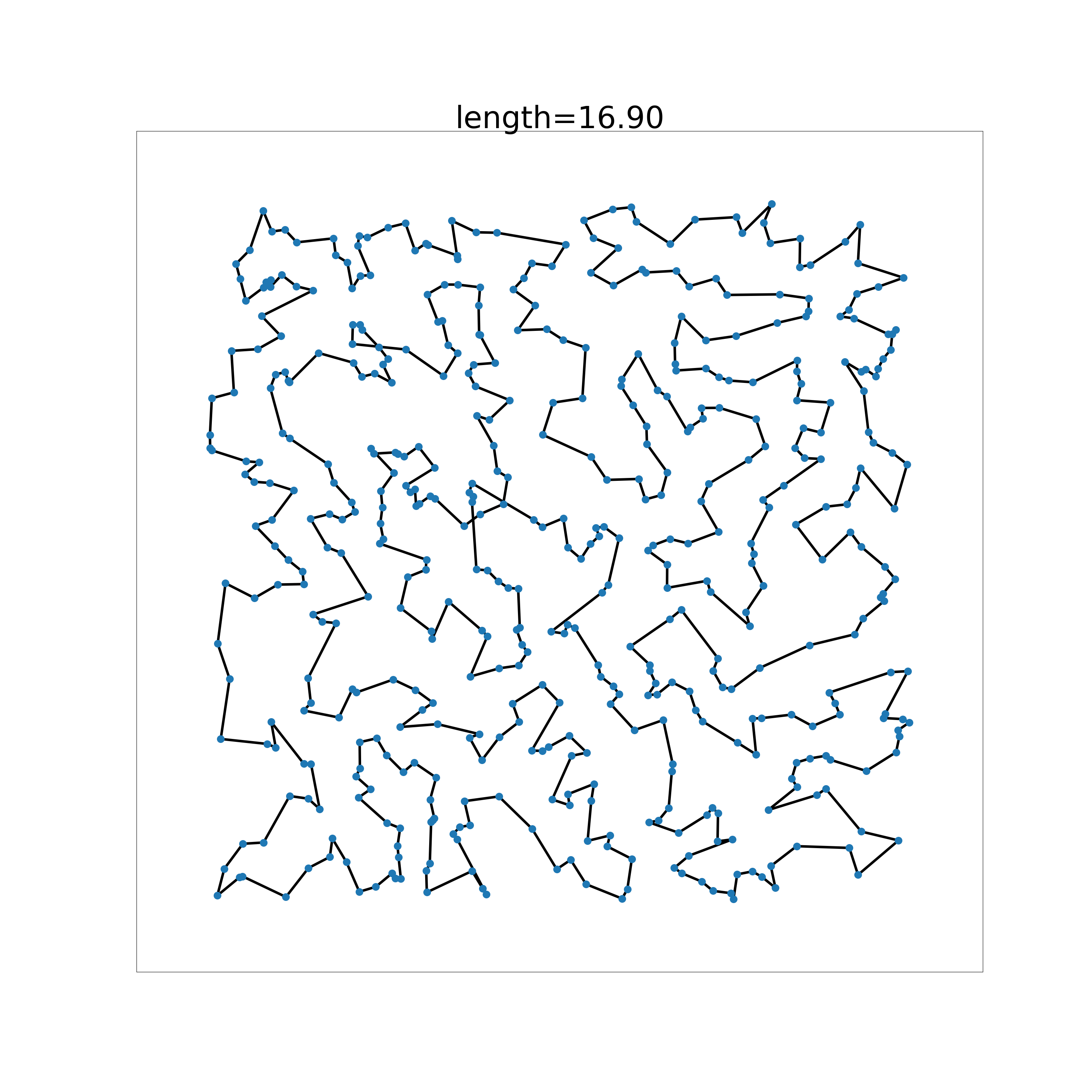}}\quad
	\subfloat[TSP\_random1000]{\includegraphics[width=0.225\textwidth]{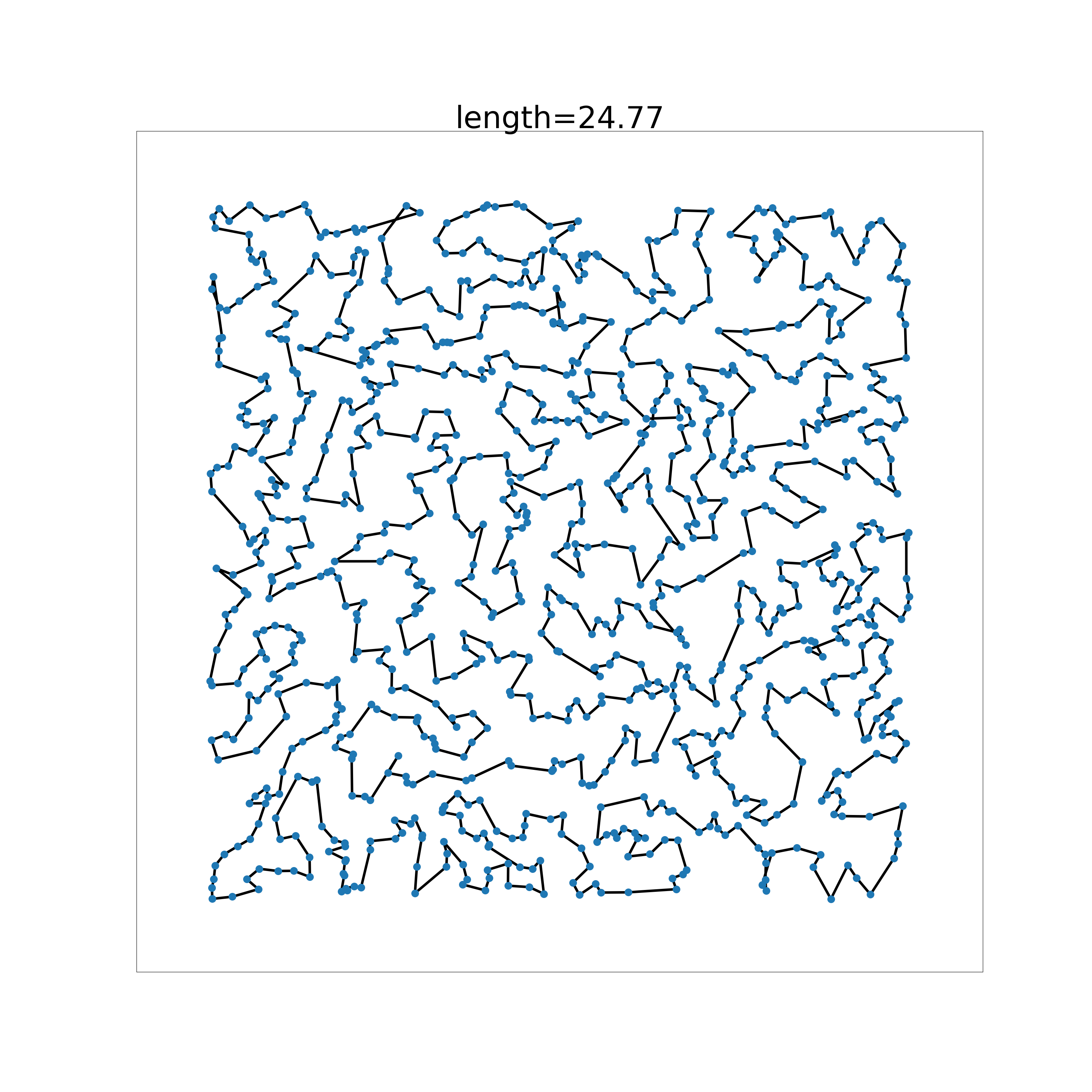}}\quad
	
	\subfloat[TSPLIB\_ts225]{\includegraphics[width=0.225\textwidth]{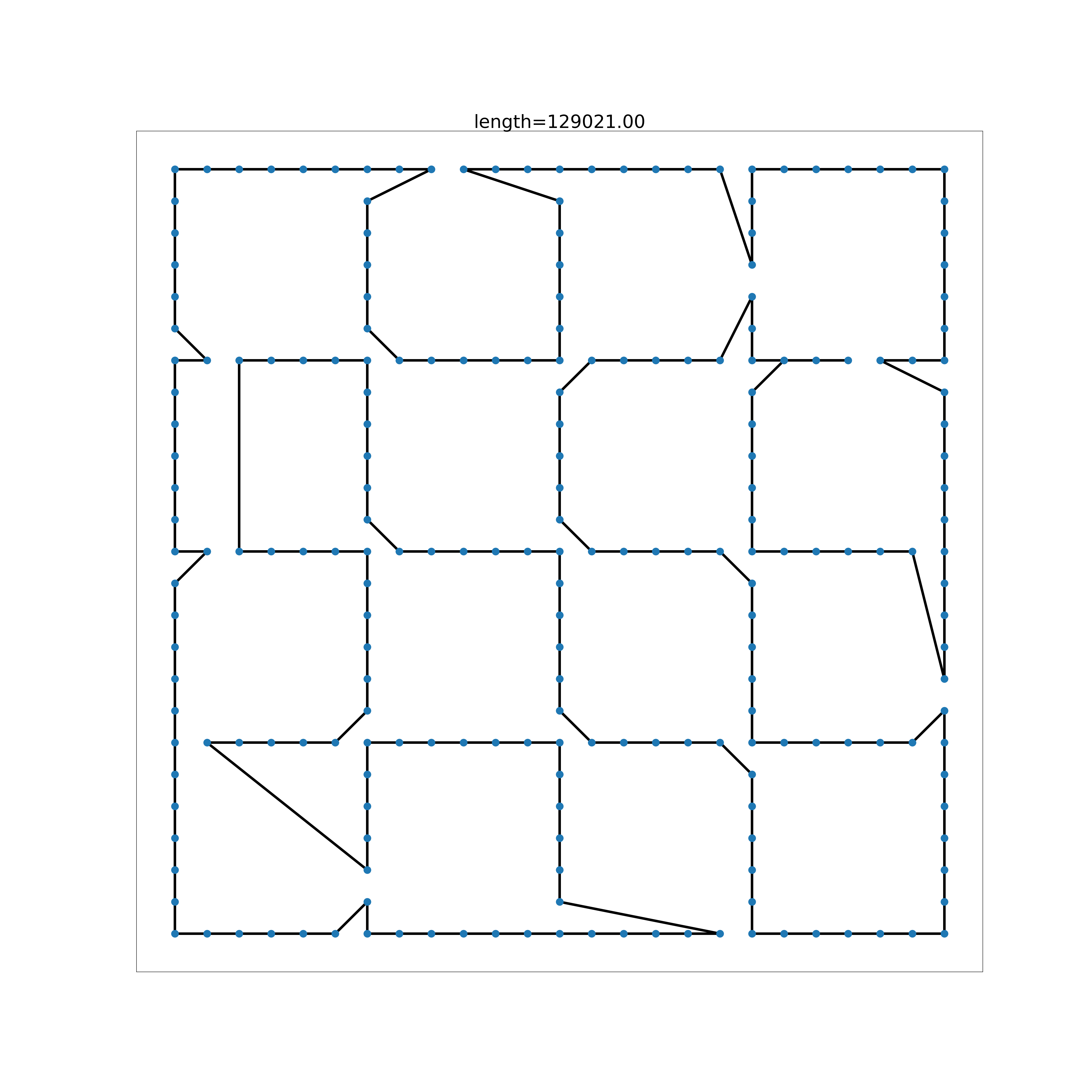}}\quad
	\subfloat[TSPLIB\_fl417]{\includegraphics[width=0.225\textwidth]{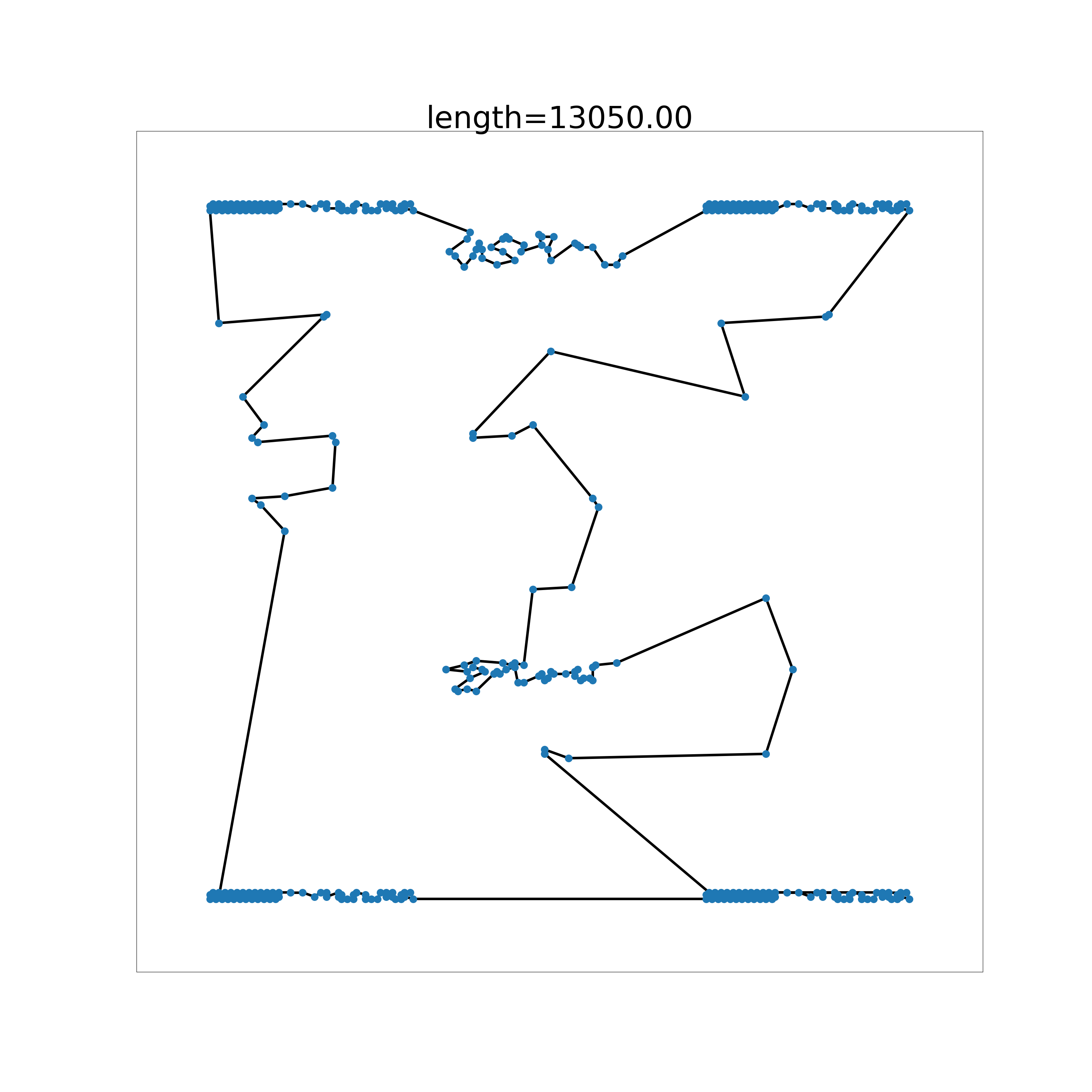}}\\
	
	\caption{Example solutions by \name. }
    \label{fig:route_visual}
\end{figure}

We list all the detailed computational results as follows. Furthermore, we also provide four example solutions obtained by our \name in Figure~\ref{fig:route_visual}. And one easily observes that even distribution of nodes in \textbf{TSPLIB} instances is different from \textbf{TSP\_random}, \name is able to generalize to different scenarios and makes good routing decisions accordingly.

\begin{table*}[!htb]
\renewcommand\arraystretch{1.1}
\centering
\caption{Detailed results on practical instances from \textbf{TSPLIB}.}
\resizebox{\textwidth}{!}{
\begin{tabular}{cccccccccccccc}
\toprule[2pt]
         & OPT    & \multicolumn{2}{c}{\name(Model100)} & \multicolumn{2}{c}{\name(Model200)} & \multicolumn{2}{c}{POMO}          & \multicolumn{2}{c}{AM}      & \multicolumn{2}{c}{DRL+2opt}                     & \multicolumn{2}{c}{AM+LCP} \\ \hline
         &        & Len.(Gap)                        & Time(s)          & Len.(Gap)                        & Time(s)         & Len.(Gap)               & Time(s) & Len.(Gap)         & Time(s) & Len.(Gap)                              & Time(s) & Len.(Gap)        & Time(s) \\ \hline
eil51    & 426    & 428(0.47\%)                      & 0.20             & \textbf{426(0.00\%)}             & 0.20            & 431(1.17\%)             & 1.33    & 435(2.11\%)       & 0.86    & 431(1.17\%)                            & 697.24  & 429(0.73\%)      & 13      \\
berlin52 & 7542   & \textbf{7542(0.00\%)}            & 0.19             & 7740(2.63\%)                     & 0.19            & 7542(0.00\%)            & 1.38    & 11336(50.30\%)    & 0.09    & 7547(0.07\%)                           & 703.77  & 7550(0.10\%)     & 13      \\
st70     & 675    & \textbf{675(0.00\%)}             & 0.17             & 678(0.44\%)                      & 0.17            & 675(0.00\%)             & 1.43    & 684(1.33\%)       & 0.12    & 681(0.89\%)                            & 868.52  & 680(0.74\%)      & 13      \\
eil76    & 538    & \textbf{538(0.00\%)}             & 0.17             & 541(0.56\%)                      & 0.17            & 538(0.00\%)             & 1.47    & 550(2.23\%)       & 0.13    & 549(2.04\%)                            & 913.19  & 547(1.64\%)      & 18      \\
pr76     & 108159 & 109117(0.89\%)                   & 0.17             & 111173(2.79\%)                   & 0.17            & \textbf{108159(0.00\%)} & 1.48    & 115300(6.60\%)    & 0.14    & 109300(1.05\%)                         & 873.01  & 108633(0.44\%)   & 18      \\
rat99    & 1211   & 1258(3.88\%)                     & 0.19             & 1274(5.20\%)                     & 0.19            & 1253(3.47\%)            & 1.38    & 1363(12.55\%)     & 0.20    & \textbf{1252(3.39\%)}                  & 1007.35 & 1292(6.67\%)     & 24      \\
kroA100  & 21282  & \textbf{21452(0.80\%)}           & 0.20             & 21817(2.51\%)                    & 0.20            & 21550(1.26\%)           & 1.41    & 28464(33.75\%)    & 0.21    & 22517(5.80\%)                          & 959.44  & 21910(2.95\%)    & 26      \\
kroB100  & 22141  & 22579(1.98\%)                    & 0.27             & 23029(4.01\%)                    & 0.27            & \textbf{22199(0.26\%)}  & 1.38    & 27755(25.36\%)    & 0.21    & 23515(6.21\%)                          & 985.26  & 22476(1.51\%)    & 26      \\
kroC100  & 20749  & 20958(1.01\%)                    & 0.20             & 21755(4.85\%)                    & 0.20            & 21017(1.29\%)           & 1.43    & 23653(14.00\%)    & 0.20    & \textbf{20856(0.52\%)}                 & 967.05  & 21337(2.84\%)    & 26      \\
kroD100  & 20749  & 21923(5.66\%)                    & 0.19             & 22546(8.66\%)                    & 0.19            & \textbf{21893(5.51\%)}  & 1.50    & 26274(26.63\%)    & 0.21    & 22045(6.25\%)                          & 949.33  & 21714(1.97\%)    & 26      \\
kroE100  & 22068  & 22362(1.33\%)                    & 0.19             & 22694(2.84\%)                    & 0.19            & 22377(1.40\%)           & 1.30    & 23415(6.10\%)     & 0.21    & \textbf{22352(1.29\%)}                 & 972.64  & 22488(1.90\%)    & 26      \\
rd100    & 7910   & \textbf{7910(0.00\%)}            & 0.20             & 7947(0.47\%)                     & 0.20            & \textbf{7910(0.00\%)}   & 1.39    & 8175(3.35\%)      & 0.20    & 7953(0.54\%)                           & 1044.88 & 22488(1.90\%)    & 26      \\
eil101   & 629    & \textbf{630(0.16\%)}             & 0.19             & 632(0.48\%)                      & 0.19            & \textbf{630(0.16\%)}    & 1.30    & 643(2.23\%)       & 0.21    & 639(1.59\%)                            & 1046.10 & 645(2.59\%)      & 26      \\
lin105   & 14379  & 14774(2.75\%)                    & 0.20             & 15545(8.11\%)                    & 0.20            & \textbf{14580(1.40\%)}  & 1.37    & 32024(122.71\%)   & 0.22    & 16161(12.39\%)                         & 1047.25 & 14934(3.86\%)    & 26      \\
pr107    & 44303  & 45307(2.27\%)                    & 0.22             & 45573(2.87\%)                    & 0.22            & \textbf{44907(1.36\%)}  & 1.42    & 46763(5.55\%)     & 0.28    & 51180(15.52\%)                         & 1009.91 & -(-\%)           & -       \\
pr124    & 59030  & \textbf{59076(0.08\%)}           & 0.21             & 59416(0.65\%)                    & 0.21            & \textbf{59076(0.08\%)}  & 1.52    & 61323(3.88\%)     & 0.28    & 59850(1.39\%)                          & 1072.35 & 61294(3.84\%)    & 37      \\
bier127  & 118282 & \textbf{121047(2.34\%)}          & 0.21             & 130513(10.34\%)                  & 0.21            & 124531(5.28\%)          & 1.41    & 135357(14.44\%)   & 0.29    & 124580(5.32\%)                         & 1154.38 & 128832(8.92\%)   & 37      \\
ch130    & 6110   & \textbf{6116(0.10\%)}            & 0.25             & 6139(0.47\%)                     & 0.25            & 6124(0.23\%)            & 1.44    & 6331(3.62\%)      & 0.30    & 6205(1.55\%)                           & 1137.89 & 6145(0.57\%)     & 38      \\
pr136    & 96772  & 97718(0.98\%)                    & 0.23             & \textbf{97466(0.72\%)}           & 0.23            & 97493(0.75\%)           & 1.39    & 102347(5.76\%)    & 0.34    & 99582(2.90\%)                          & 1193.28 & 98285(1.56\%)    & 38      \\
pr144    & 58537  & 58634(0.17\%)                    & 0.25             & \textbf{58617(0.14\%)}           & 0.25            & 58924(0.66\%)           & 1.43    & 64399(10.01\%)    & 0.36    & 60511(3.37\%)                          & 1066.46 & 60571(3.47\%)    & 43      \\
ch150    & 6528   & 6562(0.52\%)                     & 0.24             & 6570(0.64\%)                     & 0.24            & \textbf{6558(0.46\%)}   & 1.52    & 6839(4.76\%)      & 0.38    & 6618(1.38\%)                           & 1314.38 & -(-\%)           & -       \\
kroA150  & 26524  & 27410(3.34\%)                    & 0.25             & 27736(4.57\%)                    & 0.25            & \textbf{26734(0.79\%)}  & 1.47    & 31602(19.14\%)    & 0.40    & 28400(7.07\%)                          & 1311.79 & 27501(3.68\%)    & 44      \\
kroB150  & 26130  & 26810(2.60\%)                    & 0.24             & 27235(4.23\%)                    & 0.24            & \textbf{26583(1.73\%)}  & 1.51    & 32043(22.63\%)    & 0.39    & 27828(6.50\%)                          & 1245.82 & 26962(3.18\%)    & 44      \\
pr152    & 73682  & 74610(1.26\%)                    & 0.23             & 74791(1.51\%)                    & 0.23            & \textbf{74382(0.95\%)}  & 1.47    & 84778(15.06\%)    & 0.39    & 78632(6.72\%)                          & 1169.23 & 75539(2.52\%)    & 44      \\
u159     & 42080  & 42495(0.99\%)                    & 0.26             & \textbf{42487(0.97\%)}           & 0.26            & 42510(1.02\%)           & 1.46    & 45681(8.56\%)     & 0.42    & 42800(1.71\%)                          & 1286.03 & 46640(10.84\%)   & 45      \\
rat195   & 2323   & \textbf{2488(7.10\%)}            & 0.31             & 2490(7.19\%)                     & 0.31            & 2549(9.73\%)            & 1.49    & 2763(18.94\%)     & 0.58    & \textbackslash{}textbf\{2455(5.68\%)\} & 1639.36 & 2574(10.81\%)    & 57      \\
d198     & 15780  & \textbf{18290(15.91\%)}          & 0.32             & 23390(48.23\%)                   & 0.32            & 18766(18.92\%)          & 1.39    & 85221(440.06\%)   & 0.60    & 20982(32.97\%)                         & 1439.20 & -(-\%)           & -       \\
kroA200  & 29368  & 30905(5.23\%)                    & 0.31             & 30509(3.89\%)                    & 0.31            & \textbf{29939(1.94\%)}  & 1.40    & 38739(31.91\%)    & 0.59    & 32208(9.67\%)                          & 1558.14 & 31172(6.14\%)    & 86      \\
kroB200  & 29437  & 30978(5.23\%)                    & 0.31             & 30578(3.88\%)                    & 0.31            & \textbf{30522(3.69\%)}  & 1.42    & 38106(29.45\%)    & 0.62    & 31422(6.74\%)                          & 1549.52 & -(-\%)           & -       \\
ts225    & 126643 & 134522(6.22\%)                   & 0.35             & \textbf{128755(1.67\%)}          & 0.35            & 136221(7.56\%)          & 1.58    & 140768(11.15\%)   & 0.71    & 132677(4.76\%)                         & 1622.01 & 134827(6.46\%)   & 113     \\
tsp225   & 3916   & 4269(9.01\%)                     & 0.37             & 4233(8.09\%)                     & 0.37            & \textbf{4142(5.77\%)}   & 1.54    & 5249(34.04\%)     & 0.71    & 4224(7.87\%)                           & 1766.53 & 4487(14.50\%)    & 113     \\
pr226    & 80369  & \textbf{81593(1.52\%)}           & 0.34             & 82275(2.37\%)                    & 0.34            & 82632(2.82\%)           & 1.51    & 93144(15.90\%)    & 0.72    & 90017(12.00\%)                         & 1485.56 & 85262(6.09\%)    & 113     \\
gil262   & 2378   & \textbf{2425(1.98\%)}            & 0.43             & 2409(1.30\%)                     & 0.43            & 2440(2.61\%)            & 1.56    & 2696(13.37\%)     & 0.92    & 2476(4.12\%)                           & 1973.80 & 2508(5.49\%)     & 134     \\
pr264    & 49135  & \textbf{51494(4.80\%)}           & 0.45             & 52489(6.83\%)                    & 0.45            & 55003(11.94\%)          & 1.59    & 68677(39.77\%)    & 0.94    & 66737(35.82\%)                         & 1856.47 & -(-\%)           & -       \\
a280     & 2579   & 2824(9.50\%)                     & 0.51             & \textbf{2724(5.62\%)}            & 0.51            & 2837(10.00\%)           & 1.67    & 3330(29.12\%)     & 1.03    & 2808(8.88\%)                           & 2007.79 & -(-\%)           & -       \\
pr299    & 48191  & 54718(13.54\%)                   & 0.54             & 54407(12.90\%)                   & 0.54            & \textbf{53754(11.54\%)} & 1.60    & 256687(432.65\%)  & 1.16    & 61145(26.88\%)                         & 2159.60 & -(-\%)           & -       \\
lin318   & 42029  & 44312(5.43\%)                    & 0.62             & \textbf{44117(4.97\%)}           & 0.62            & 45423(8.08\%)           & 1.58    & 51507(22.55\%)    & 1.29    & 47456(12.91\%)                         & 2236.29 & 46540(10.72\%)   & 158     \\
rd400    & 15281  & 16659(9.02\%)                    & 0.99             & \textbf{15848(3.71\%)}           & 0.99            & 17056(11.62\%)          & 1.75    & 19010(24.40\%)    & 1.93    & 17010(11.31\%)                         & 2804.61 & 16519(8.10\%)    & 209     \\
f1417    & 11861  & \textbf{12782(7.76\%)}           & 1.05             & 12860(8.42\%)                    & 1.05            & 13529(14.06\%)          & 1.83    & 25813(117.63\%)   & 2.09    & 17655(48.85\%)                         & 2623.33 & -(-\%)           & -       \\
pr439    & 107217 & 121515(13.34\%)                  & 1.18             & \textbf{117302(9.41\%)}          & 1.18            & 122512(14.27\%)         & 1.96    & 386134(260.14\%)  & 2.39    & 151400(41.21\%)                        & 2796.83 & 130996(22.18\%)  & 228     \\
pcb442   & 50778  & 56432(11.13\%)                   & 1.22             & \textbf{53927(6.20\%)}           & 1.22            & 58287(14.79\%)          & 1.84    & 67828(33.58\%)    & 2.42    & 59257(16.70\%)                         & 2638.25 & 57051(12.35\%)   & 228     \\
d493     & 35002  & \textbf{41501(18.57\%)}          & 1.56             & 63287(80.81\%)                   & 1.56            & 50939(45.53\%)          & 2.00    & 228336(552.35\%)  & 2.92    & 56627(61.78\%)                         & 3050.07 & -(-\%)           & -       \\
u574     & 36905  & 43517(17.92\%)                   & 2.40             & 42967(16.43\%)                   & 2.40            & 45356(22.90\%)          & 2.35    & 75761(105.29\%)   & 3.73    & \textbf{40197(8.92\%)}                 & 3838.43 & -(-\%)           & -       \\
rat575   & 6773   & 7991(17.98\%)                    & 2.40             & \textbf{7844(15.81\%)}           & 2.40            & 8322(22.87\%)           & 2.36    & 10388(53.37\%)    & 3.75    & 8539(26.07\%)                          & 4108.52 & -(-\%)           & -       \\
p654     & 34643  & \textbf{38463(11.03\%)}          & 3.53             & 39138(12.98\%)                   & 3.53            & 41978(21.17\%)          & 2.91    & 117482(239.12\%)  & 4.77    & 55372(59.84\%)                         & 3200.09 & -(-\%)           & -       \\
d657     & 48912  & \textbf{58113(18.81\%)}          & 3.61             & 63151(29.11\%)                   & 3.61            & 64789(32.46\%)          & 2.99    & 208309(325.89\%)  & 4.85    & 78444(60.38\%)                         & 4102.30 & -(-\%)           & -       \\
u724     & 41910  & 50497(20.49\%)                   & 4.88             & \textbf{48292(15.23\%)}          & 4.88            & 52849(26.10\%)          & 3.44    & 79541(89.79\%)    & 5.79    & 63492(51.50\%)                         & 4751.20 & -(-\%)           & -       \\
rat783   & 8806   & 10764(22.23\%)                   & 6.22             & \textbf{10562(19.94\%)}          & 6.22            & 11242(27.66\%)          & 3.96    & 15432(75.24\%)    & 6.68    & 14815(68.24\%)                         & 5592.80 & -(-\%)           & -       \\
pr1002   & 259045 & 316225(22.07\%)                  & 12.91            & \textbf{305455(17.92\%)}         & 12.91           & 350600(35.34\%)         & 6.43    & 477737(84.42\%)   & 10.97   & 318725(23.04\%)                        & 5674.53 & -(-\%)           & -       \\ \bottomrule[2pt]
\end{tabular}
}
\end{table*}

\end{document}